\documentclass{bmvc2k}
\usepackage{cleveref}
\usepackage{amsmath}
\usepackage{amssymb}
\usepackage{booktabs}
\usepackage{colortbl}

\title{M$^2$StyleGS: Multi-Modality 3D Style Transfer with Gaussian Splatting}

\addauthor{Xingyu Miao}{xingyu.miao@durham.ac.uk}{1,*}
\addauthor{Xueqi Qiu}{xueqi.qiu@durham.ac.uk}{1,*}
\addauthor{Haoran Duan}{haoranduan@tsinghua.edu.cn}{2}
\addauthor{Yawen Huang}{yawenhuang@tencent.com}{3}
\addauthor{Xian Wu}{kevinxwu@tencent.com}{3}
\addauthor{Jingjing Deng}{jingjing.deng@bristol.ac.uk}{4,$\dagger$}
\addauthor{Yang Long}{yang.long@durham.ac.uk}{1,$\dagger$}
\addinstitution{
 Department of Computer Science\\
 Durham University\\
 Durham, UK
}
\addinstitution{
Department of Automation\\
Tsinghua University\\ 
Beijing, China
}
\addinstitution{
 Tencent Jarvis Lab\\
 Shenzhen, China
}
 \addinstitution{
 School of Engineering Mathematics and Technology\\
 Bristol University,
 Bristol, UK
}

\runninghead{Miao, Qiu, Duan, Huang, Wu, Deng, Long}{M$^2$StyleGS}

\begin{document}

\maketitle
\begin{abstract}
Conventional 3D style transfer methods rely on a fixed reference image to apply artistic patterns to 3D scenes. However, in practical applications such as virtual or augmented reality, users often prefer more flexible inputs, including textual descriptions and diverse imagery. In this work, we introduce a novel real-time styling technique M$^2$StyleGS to generate a sequence of precisely color-mapped views. It utilizes 3D Gaussian Splatting (3DGS) as a 3D presentation and multi-modality knowledge refined by CLIP as a reference style. M$^2$StyleGS resolves the abnormal transformation issue by employing a precise feature alignment, namely ``subdivisive flow”, it strengthens the projection of the mapped CLIP text-visual combination feature to the VGG style feature. In addition, we introduce observation loss, which assists in the stylized scene better matching the reference style during the generation, and suppression loss, which suppresses the offset of reference color information throughout the decoding process. By integrating these approaches, M$^2$StyleGS can employ text or images as references to generate a set of style-enhanced novel views. Our experiments show that M$^2$StyleGS achieves better visual quality and surpasses the previous work by up to 32.92\% in terms of consistency. For more visualization results and code, please refer to the project page: https://nora202.github.io/MMStyleGS/
\end{abstract}

\section{Introduction}
\label{sec:intro}
Style transfer has been a fascinating interdisciplinary research in computer vision. Transferring the artistic style in 2D painting is the predominant direction at the early stage, where \cite{gatys2016image,huang2017arbitrary,sheng2018avatar,li2019learning,park2019arbitrary,deng2020arbitrary,liu2021adaattn,wu2021styleformer} create visually stunning new images. After that, style transfer approaches were extended to video and dynamic content \cite{chen2017coherent,huang2017real,gao2019reconet,chen2020optical,wang2020consistent}.

With the development of 3D data acquisition, virtual reality and augmented reality, style transfer has been gradually adopted in various 3D applications \cite{chiang2022stylizing,fan2022unified,huang2022stylizednerf,nguyen2022snerf,zhang2022arf,liu2023stylerf,chen2024upst,miao2024conrf}. In 3D style transfer, the advanced methods are divided into two categories: implicit and explicit. Implicit transformation edits the rendered images through Score Distillation Sampling (SDS) loss \cite{poole2022dreamfusion} and adjusts the 3D space parameter through backpropagation. Explicit transformation utilizes neural networks \cite{liu2024stylegaussian,saroha2024gaussian} to extract features of reference images, and applies them to target 3D scenes with style transfer modules, such as AdaIN \cite{liu2021adaattn,wu2022ccpl}. Compared to implicit methods, explicit methods show unique advantages in modeling the transformation between reference and target features \cite{chen2017stylebank,mehra2021implicit}, among which VGG \cite{simonyan2014very} can better capture the subtle semantic differences between artistic styles than other neural networks \cite{wang2021rethinking}. Therefore, our proposed M$^2$StyleGS focuses on VGG-based explicit transformation.

Existing 3D style transformation request a reference style image \cite{liu2023stylerf,liu2024stylegaussian,huang2022stylizednerf}, 
 while many users also look forward to transferring the style through diversity description. To resolve this limitation, ConRF \cite{miao2024conrf} leverages the CLIP text-visual pair feature with an MLP network to synchronize the CLIP text-visual feature to the VGG style feature. Nevertheless, due to the ambiguity and diversity of CLIP feature distribution, we found that ConRF just implements rough alignment. As a result, it transfers the abnormal colors and blurred textures, and inconsistencies exist between the style image and the corresponding textual description.
\begin{figure*}[t]
    \centering
    \includegraphics[width=1\textwidth]{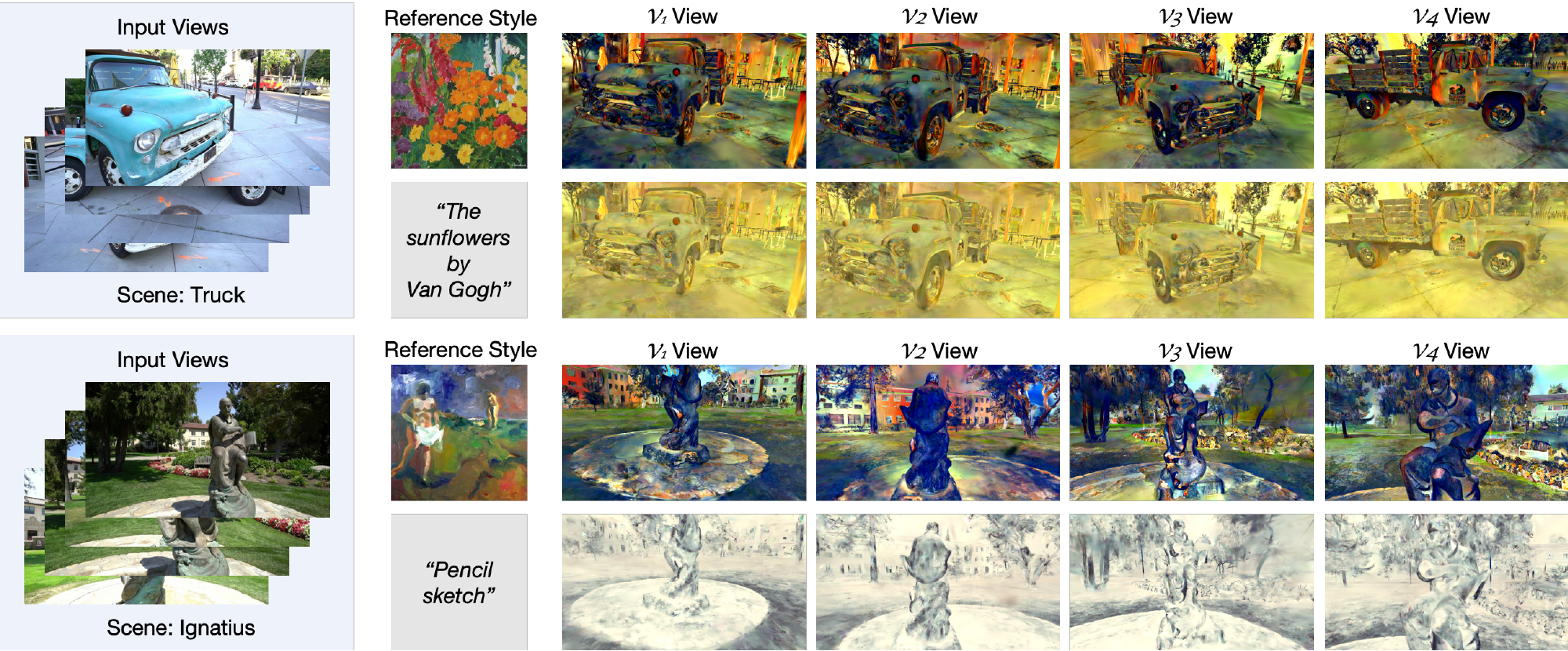}
    \caption{\textbf{Multi-modality 3D style transfer with M$^2$StyleGS.} Employing a set of 3D scene images captured from various perspectives, M$^2$StyleGS can effectively apply reference styles described in arbitrary images or text. }
    \label{examples}
\end{figure*}

In this work, we aim to overcome the color anomalies and smoothed textures issues, thereby achieving high-quality multi-modality 3D-based style transfer to meet user descriptions and expectations. Specifically, we introduce a novel feature alignment strategy to alleviate style transfer problems caused by inconsistent feature distribution, which we term ``subdivisive flow”. This approach finely tunes the flow direction of alignment between the mapped CLIP text-visual features with the VGG-based style feature space. To further enhance stylization performance while maintaining multi-view consistency, we also introduce auxiliary losses, namely observation loss and suppression loss, to regularize the transfer process across the global 3D scene. The contributions of our work are summarized as follows:
\begin{itemize} 
  \item We investigate the existing transformation issues such as abnormal color generated in previous work, and conclude that this is due to distinct feature distributions with a rough artistic feature domain transfer. 
  
  \item We propose a novel flow-matching method that employs the ordinary differential equation (ODE) function to enable a more precise alignment of CLIP text-visual pair features with the VGG style feature. This method not only connects the source and target features but also incorporates all intermediate features generated during the alignment.
  
  \item We introduce two auxiliary loss terms to enhance the visual effect: observation loss and suppression loss. The former optimizes the generation process to make the stylized 3D scene closer to the pre-trained 2D stylized image as observation priori, while the latter modifies the offset of the transferred color distribution at the scene-level.
  
  \item We achieve state-of-the-art (SOTA) performance by integrating explicit multi-modal style transfer with 3DGS, surpassing single-input style transfer methods in both long-term and short-term consistency.
\end{itemize}

\begin{figure*}[]
    \centering
    \includegraphics[width=1\textwidth]{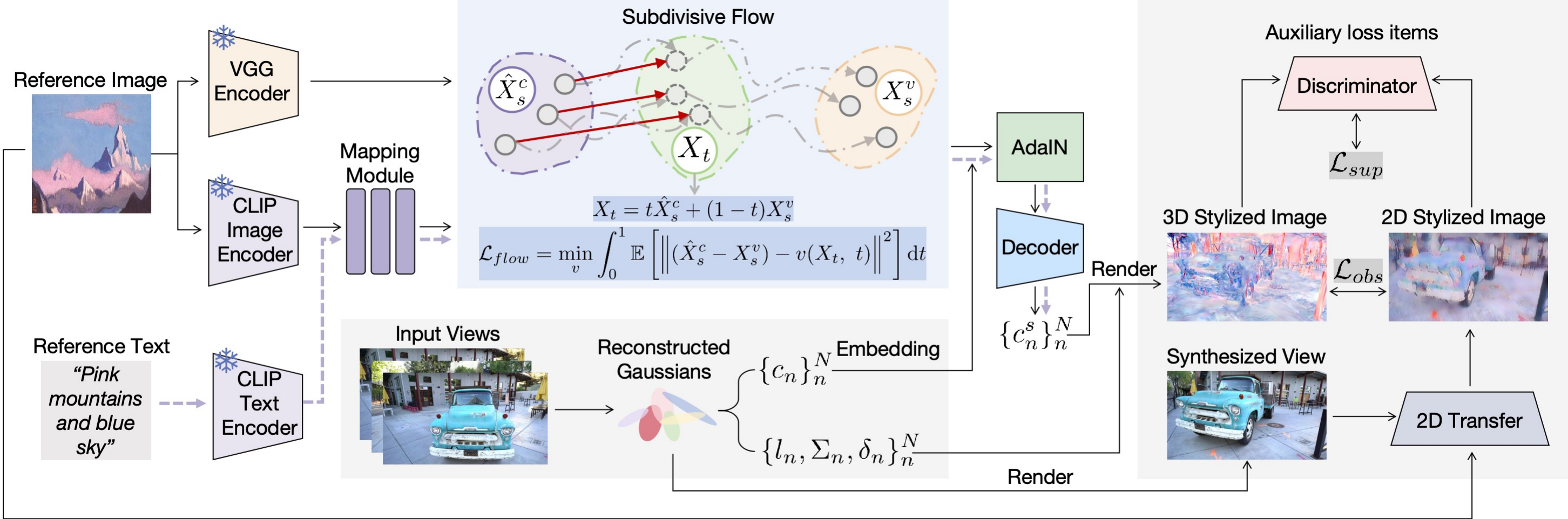}
    \caption{\textbf{The pipeline of M$^2$StyleGS.} During the training phase, M$^2$StyleGS performs style transfer using style images. After training, M$^2$StyleGS is capable of applying multi-modality stylistic transformations directly using either images or text (the purple dashed arrows illustrate text feature processing.)}
    \label{pipeline}
\end{figure*}

\section{Related Work}
\subsection{Style Transfer on Image and Video}
The objective of artistic image stylization is to integrate the content image with the stylistic image seamlessly. The seminal work of \cite{gatys2016image} succeeded in extracting semantic information from style images using pre-trained VGG \cite{simonyan2014very} to implement 2D style transfer. The different layers of the VGG network are equipped to extract various levels of artistic information from style images \cite{jing2019neural}. In addition, advancements in techniques such as optical flow \cite{dosovitskiy2015flownet} enabled researchers \cite{chen2017coherent,chen2020optical,huang2017real,gao2019reconet,wang2020consistent} to expand 2D style transfer methodologies to video sequences. After that, neural network-based style transfer has progressively advanced from transferring a single style to arbitrary styles \cite{huang2017arbitrary,sheng2018avatar,li2019learning,park2019arbitrary,deng2020arbitrary,liu2021adaattn}. Above concepts have been widely implemented in most 2D style transfers and have been extended to 3D style transfer.

\subsection{Style Transfer on 3D Scene}
NeRF \cite{mildenhall2021nerf} have initially demonstrated significant potential for high-quality 3D scene rendering. Specially, several studies \cite{chiang2022stylizing,chen2024upst,fan2022unified,huang2022stylizednerf,liu2023stylerf,nguyen2022snerf,zhang2022arf,miao2024conrf} integrate NeRF with style transfer \cite{gatys2016image}. In particular, ConRF \cite{miao2024conrf} implements multi-modality style transfer by converting the CLIP feature space into the style space of a pre-trained VGG network.

Compared to NeRF, 3DGS \cite{kerbl20233d} recently stands out for its real-time rendering performance and excellent reconstruction results. The cornerstone of its instant rendering capability relies on rasterization rather than ray tracing to render the scene. However, there is limited research combining 3DGS with artistic style transfer to deal with artistic 3D scene stylization. StyleGaussian \cite{liu2024stylegaussian} first proposes an efficient feature rendering strategy in style transfer. This method renders low-dimensional features and maps them into high-dimensional ones while embedding VGG features. GSS \cite{saroha2024gaussian} employs a pre-trained Gaussian and processes it using a multi-resolution hash grid and a tiny MLP to obtain a conditional stylized view. StylizedGS \cite{zhang2024stylizedgs} introduces a sophisticated degree of perceptual controllability, allowing users to customize functionality by manipulating colors and scales of specific zones. 

Nevertheless, these 3DGS-based methods are primarily based on single-style transfer. To our knowledge, our approach is the \textbf{first} to implement arbitrary scene stylization using multi-modal inputs with instant 3DGS as a backbone. 

\begin{figure*}[t]
    \centering
    \includegraphics[width=1\textwidth]{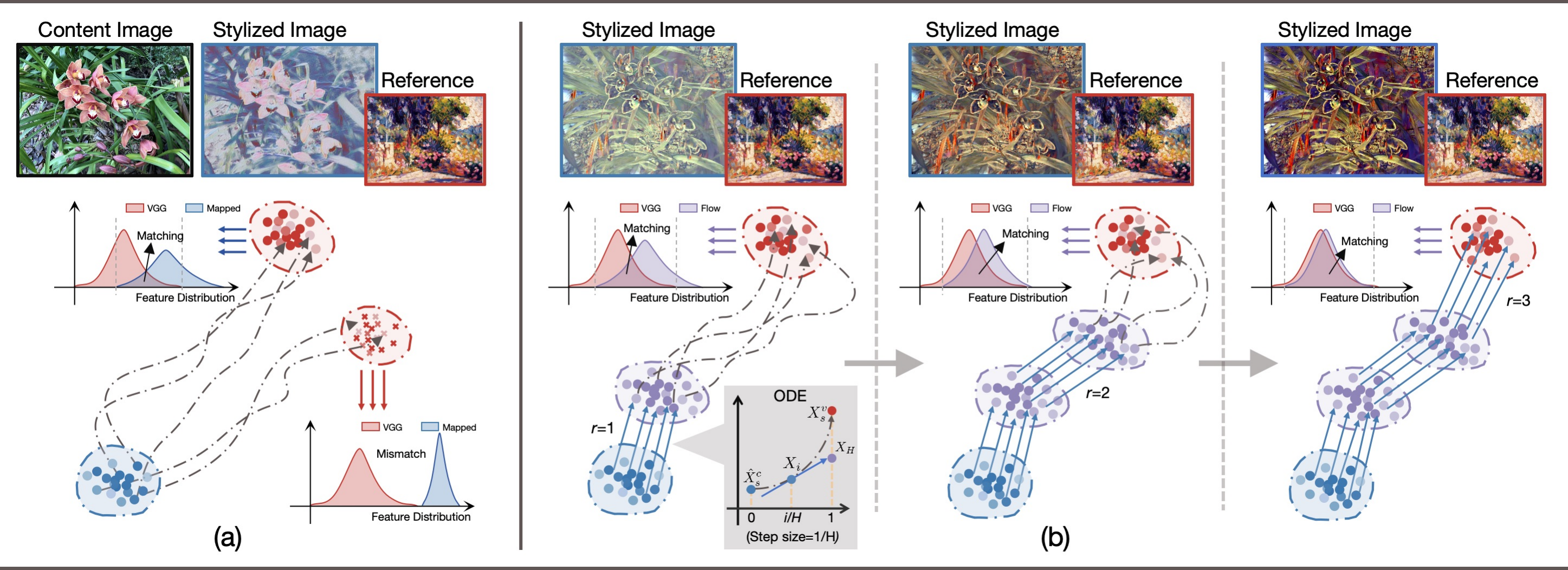}
    \caption{\textbf{Comparison of the previous alignment method with the proposed subdivisive flow.} (a) illustrates that the direct use of the mapping module can result in mismatches in feature projection. (b) illustrates the subdivisive flow learns the ODE that characterizes the trajectory from the CLIP feature flow to the VGG feature space precisely.}
    \label{motivation}
\end{figure*}

\section{Method}

\subsection{Preliminaries}

Following \cite{kerbl20233d,liu2024stylegaussian}, we represent a 3D scene $G$ as a set of 3D Gaussian primitives: 
\begin{equation}
    G=\{g_{n}=\{l_{n},\Sigma_{n},\delta _{n},c_{n}\}\}_{n}^{N},\quad n \in [1,N],
\end{equation}
where each Gaussian $g_{n}$ is parameterized by a mean $l_{n} \in \mathbb{R}^{3}$ indicating its centre location, a covariance matrix $\Sigma_{n} \in \mathbb{R}^{3\times3}$ indicating its shape and size, an opacity $\delta_{n} \in \mathbb{R}$ and color $c_{n} \in \mathbb{R}^{3}$. The goal of our work is to stylize the color of each Gaussian while maintaining original Gaussians content structure, for each Gaussian, we fix its geometry and only transform its color, and obtain stylized Gaussian $g_n^s$:
\begin{equation}
    g_{n}^{s}:\{{l_{n},\Sigma_{n},\delta _{n},c_{n}}\}\rightarrow\{{l_{n},\Sigma_{n},\delta _{n},c_{n}^{s}}\}.
    \label{aim}
    \end{equation}

\begin{figure*}[t]
    \centering
    \includegraphics[width=1\textwidth]{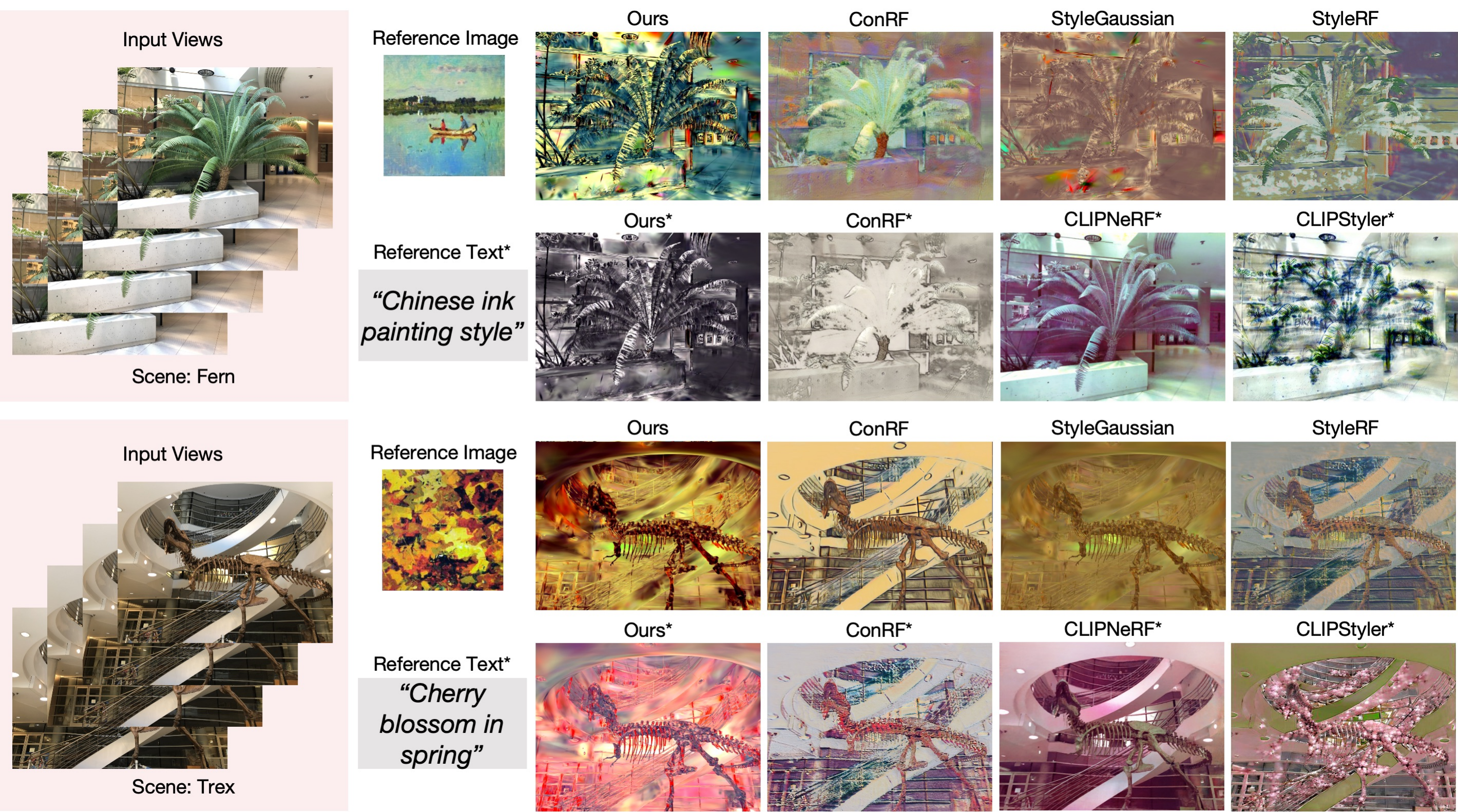}
    \caption{\textbf{Qualitative comparison with SOTA 3D style transfer methods.} $\mathrm{M^{2}StyleGS}$ can achieve precise style transfer based on the reference style images or reference style text.} 
    \label{style_img}
\end{figure*}

To achieve multi-modal transfer, as shown in \Cref{pipeline}, in M$^2$StyleGS, the CLIP branch begins with a CLIP image encoder $\mathcal{E}$, which extracts the feature $X_{s}^{c}$ from the style image $I_s$. The mapping model $\mathcal{F}_c$ establishes a correspondence between the CLIP feature vectors $X_{s}^{c}$ and in-depth style representation feature $X_s^{v}$ extracted by VGG $\mathcal{V}$:
\begin{equation}
X_s^{v}=\mathcal{V}(I_{s}),\quad  X_{s}^{c}=\mathcal{E}(I_{s}),\quad \hat{X}_{s}^{c}=\mathcal{F}_{c}(X_{s}^{c}).
    \label{previous_mapping}
\end{equation}

We embed 3D Gaussians color $\{c_n\}_n^{N}$ as $\{c_n^{e}\}_n^{N}$, and keep them fixed for subsequent stylization. Following previous researches \cite{wu2022ccpl,liu2024stylegaussian,miao2024ctnerf}, the intrinsic of artistic style can be commonly defined by the mean $\mu$ and standard deviation $\sigma$ of extracted feature $\hat{X}_s^{c}$, and we use the AdaIN transfer module \cite{huang2017arbitrary} to transfer $\hat{X}_s^{c}$ into $\{c_n^{e}\}_n^{N}$ and obtain $\{c_n^{'}\}_n^{N}$:
\begin{equation}
    c_{n}^{'}=\sigma(\hat{X}_{s}^{c})\times{\left(\frac{c_{n}^{e}-\mu({\{c_n^{e}\}_n^{N}})}{\sigma{(\{c_n^{e}\}_n^{N})}}\right)}+\mu(\hat{X}_{s}^{c}),
    \label{adain}
\end{equation}
and with the decoder $\mathcal{D}$, we can achieve the color transfer in \Cref{aim}: 
\begin{equation}
\{c_n^{s}\}_{n}^{N}=\mathcal{D}(\{c_n^{'}\}_n^{N}). 
\label{decoder_for_color}
\end{equation}

\subsection{Subdivisive Flow for Fine Projection}
The prior work \cite{miao2024conrf} also introduces a straightforward method that uses a projection network $\mathcal{F}_c$ to align the CLIP text-visual combination feature to VGG style feature to achieve multi-modality style transfer as in \Cref{previous_mapping}. However, this method is a coarse alignment, since directly mapping CLIP multi-modality space to target space adopts the MSE loss to regularize this mapping network $\mathcal{F}_c$:
\begin{equation}
    l= \left \| \hat{X}_{s}^{c} - X_s^{v} \right \| ^2,
\end{equation}
which is typically based on the assumption that features follow a Gaussian distribution to achieve the best constrain effect \cite{mathieu2015deep}.  
We investigated through previous work \cite{miao2024conrf} and found that for the style features extracted by the CLIP model, its distribution is unpredictable and multivariate, and does not always trend towards a Gaussian distribution. As shown in \Cref{motivation} (a), there is an obvious style mismatch between the generated and the reference images when using a single mapping module with MSE loss. To this end, we propose a more universally effective method to address domain alignment in style transfer.

\begin{figure*}[t]
    \centering
    \includegraphics[width=1\textwidth]{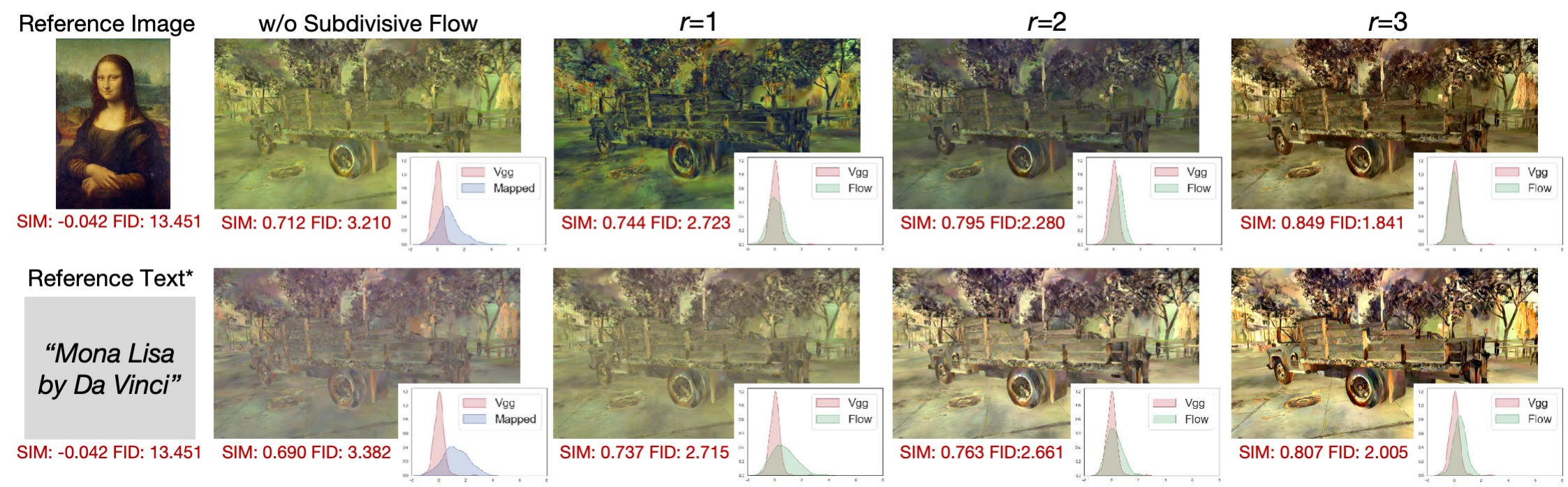}
    \caption{\textbf{Ablation studies on subdivisive flow.} The top row shows the $r$-th subdivisve round outcomes and their feature distributions among image style transfer. The bottom row shows the corresponding text style transfer, \emph{i.e.}, the original text feature distribution is the same as the style image feature distribution. Hence, the initial SIM and FID are the same.}    \label{abl}
\end{figure*}

\subsubsection{Ordinary Differential Equation (ODE)}Inspired by recent advancements in flow-based generative methods, we introduce subdivisive flow as illustrated in \Cref{motivation} (b). This formula makes the source CLIP text-visual domain stream to the target VGG domain, with a phasetically ODE lead the flow orientation. We regard the CLIP feature space $\hat{X}_s^{c}\sim \pi_{c}$ and the VGG feature space $X_{s}^{v}\sim \pi_{v}$ as two distinct domains. The flow generated from the pair ($\hat{X}_{s}^{c},X_{s}^{v}$) constitutes ODE model over time $t \in [0,1]$, which can be expressed as:
\begin{equation}
\mathrm{d}X_t = v(X_t,t)\mathrm{d}t,
\quad \text{where} \quad  t=0, X_t=\hat{X}_{s}^{c}; \quad t=1,X_t=X_{s}^{v}.
\label{new_begin}
\end{equation}

The drift force $v:\mathbb{R}^{d\times256} \to \mathbb{R}^{d\times256}$ is designed to steer the flow predominantly along the linear path from $\hat{X}_{s}^{c}$ to $X_{s}^{v}$: ($\hat{X}_{s}^{c} - X_{s}^{v}$), by the straightforward least squares regression:
\begin{equation}
\min_{v} \int_0^1\mathbb{E} \left [ \left \| ( \hat{X}_{s}^{c} - X_s^{v}) - v (X_t,~ t)\right \|^2 \right ] \mathrm{d}t, 
~~~~~\text{where}~~~~ X_t = t \hat{X}_{s}^{c} + (1-t) X_s^{v},
\label{min}
\end{equation}
$X_t$ represents the linear interpolation between $\hat{X}_{s}^{c}$ and $X_{s}^{v}$. Naively, $X_t$ adheres to the ODE given by:
\begin{equation}
    \mathrm{d}X_t =(\hat{X}_{s}^{c}-X^{v}_{s})\mathrm{d}t.
\end{equation}
Specifically, after obtain the subdivisive interpolation feature $X_t$ at random time step $t$, we use it to align with the VGG feature. This alignment ensures the distribution of all intermediate process is consistent with the VGG feature. Since the ODE is a discrete function, we simulate subdivisive flow using the Euler method for $H$ steps with step size $\triangle t=1/H$, $X_i$ is the subdivisive feature at $i$-th step, the minimization problem can be formulated as:
\begin{equation}
\min_{v} \sum_0^1\,\mathbb{E} \left[ \left\| (X_{i} - x_s^{v}) - v (F_t,~ t)\right\|^2 \right] \, \triangle t, \quad 
\text{where} \quad F_t = t F_{i} + (1-t) F_s^{v}.
\label{flow}
\end{equation}

After sufficient training to ensure that the ensemble of $t$ covers $[0, 1]$, as we gradually increase $i$, theoretically, at $i=H$, the endpoint feature distribution $X_{H}$ of the entire flow precisely aligns with that of the VGG feature distribution. 
\subsubsection{Flow Segmentation}Moreover, as illustrated in \Cref{motivation} (b), the subdivisive flow can be divided into $r$ rounds to refine the alignment process by utilizing multiple interpolations as new starting points. Except for the initial round that set $X_{s}^{v}$ as a beginning point, in each subsequent round, we update the beginning point indicated in \Cref{new_begin} with $X_{H}$ calculated from the previous round, i.e., $t=0, X_t=X_H$. Consequently, the projected feature $X_{H}$ obtained from \Cref{flow} can be aligned more effectively with the VGG feature. This approach segments the domain transformation into phased route tasks, thereby refining the alignment more efficiently.

\begin{table}[t]
\centering
\resizebox{\linewidth}{!}{
\begin{tabular}{ccccc|ccccc}
\toprule
Image reference       & \multicolumn{2}{c}{Long-range} & \multicolumn{2}{c|}{Short-range}  & Text reference       & \multicolumn{2}{c}{Long-range} & \multicolumn{2}{c}{Short-range}  \\ \cmidrule(lr){1-1} \cmidrule(lr){2-3} \cmidrule(lr){4-5} \cmidrule(lr){6-10} 
 & \textbf{LPIPS}  
              &\textbf{RMSE}                                      & \textbf{LPIPS}                                     & \textbf{RMSE}&   & \textbf{LPIPS}  
              &\textbf{RMSE}                                      & \textbf{LPIPS}                                     & \textbf{RMSE}                                                                     \\
StyleRF &  0.142      & 0.120&      0.074      &   0.073  & $\mathrm{CLIPStyler}$&0.215&  0.155 & 0.143 &  0.152  \\
StyleGaussian &  \underline{0.107}   &  0.114& \underline{0.061}   &    \underline{0.067}  & $\mathrm{CLIPNeRF}$ & \underline{0.131} &  \underline{0.106}  &  0.088 & \underline{0.079} \\
ConRF  &  0.123     &    \underline{0.100}    &     0.082 & 0.070 & $\mathrm{ConRF}$ &0.144   &  0.128   & \underline{0.075}& 0.080   \\ 

\rowcolor{pink!40}{Ours}  &  {\textbf{0.099}}  & \textbf{0.089}& \textbf{0.055} & \textbf{0.052} &{$\mathrm{Ours}$} & \textbf{0.120} & \textbf{0.102} & \textbf{0.070}& \textbf{0.057} \\ \bottomrule              
\end{tabular}
}
\caption{\textbf{Consistency results.} We compared $\mathrm{M^{2}StyleGS}$ with the SOTA methods on consistency using RMSE($\downarrow$) and LPIPS($\downarrow$) within the LLFF dataset. We employed adjacent views to obtain short-range consistency and distant views to obtain long-range consistency. Note that in all the tables, the best score is in \textbf{bold}, and the second score is \underline{underline}.}
\label{quan}
\end{table}

\subsection{Enhanced Transfer Auxiliary Loss}

\subsubsection{Observation Loss} We introduce the observation loss $\mathcal{L}_{obs}$ to enable the training of transferred color to align more closely with the reference color in holistic observation. We integrate a pre-trained 2D style transfer model as a generator $\mathcal{G}$. This model uses the content image $I_{c}$ and style image $I_s$ to produce the synthesized 2D style transfer image $I_{g} = \mathcal{G}(I_{c}, I_{s})$. $I_{g}$ is served as an observational prior, which we consider as authentic artwork. M$^2$StyleGS synthesizes the counterfeit artwork $I_{f}$ after using \Cref{adain} and \Cref{decoder_for_color}, and $\mathcal{L}_{obs}$ measures the differences between $I_{g}$ and $I_{f}$, $\phi_{i}$ denotes the $i-$layer in VGG network:

\begin{equation}
    \mathcal{L}_{obs}=\sum_{i=1}^{N} \left\| \phi_{i}(I_{g})-\phi_{i}(I_{f}) \right\|^{2}.
\end{equation}

We denote the loss in \Cref{flow} as $\mathcal{L}_{flow}$, and train the M$^2$StyleGS with the content loss $\mathcal{L}_{content}$ and style loss $\mathcal{L}_{style}$ in previous works \cite{huang2017arbitrary,miao2024conrf}.The overall loss function $\mathcal{L}_{stylized}$ for training the scene content field as:
\begin{equation}
    \mathcal{L}_{stylized}=\mathcal{L}_{content}+\lambda_{style}{\mathcal{L}_{style}}+
    \lambda_{obs}{\mathcal{L}_{obs}}+\lambda_{flow}{\mathcal{L}_{flow}},
\label{loss_all}
\end{equation}
where $\lambda_{*}$ control the balance between stylization effects and content preservation. 

\subsubsection{Suppression Loss.}To further train the shared-weight decoder $\mathcal{D}$ in the decoding step with the reference style, we introduce the suppression loss $\mathcal{L}_{sup}$, which serves to global suppress misleading color information. We differentiate between the synthesized fake artwork $I_{f}$ and authentic 2D stylized artwork $I_{g}$ with the multi-scale discriminator $\eta$ to obtain the suppression loss: 
\begin{equation}
    \mathcal{L}_{sup}=\mathbb{E} \left[log(\eta(I_{g})]+\mathbb{E}[log(1-\eta(I_{f}))) \right].
\end{equation}

 Utilizing these two losses, M$^2$StyleGS optimizes the transfer quality from a global 3D scene scale without interfering with the render process, hence also ensuring the multi-view coherence within the 3D scene.

\begin{table}[t]
\centering
\resizebox{\linewidth}{!}{
\begin{tabular}{ccccc|ccccc}
\toprule
Image reference      & \multicolumn{2}{c}{Long-range} & \multicolumn{2}{c|}{Short-range} & Text reference       & \multicolumn{2}{c}{Long-range} & \multicolumn{2}{c}{Short-range} \\ \midrule
 & \textbf{LPIPS}  
              &\textbf{RMSE}                                      & \textbf{LPIPS}                                     & \textbf{RMSE}   && \textbf{LPIPS}  
              &\textbf{RMSE}                                      & \textbf{LPIPS}                                     & \textbf{RMSE}                                                          \\          
{$\mathrm{Ours}$ w/o $\mathcal{L}_{obs}$ \& $\mathcal{L}_{sup}$}   &   0.112      &     0.101      & 0.076 &0.067  &{$\mathrm{Ours}$ w/o $\mathcal{L}_{obs}$ \&  $\mathcal{L}_{sup}$}  &0.127  & 0.136 &0.077 &0.065 \\ 
{$\mathrm{Ours}$ w/o} $\mathcal{L}_{obs}$        & 0.109 &  0.097    & \underline{0.062} &  \underline{0.056} & {$\mathrm{Ours}$ w/o $\mathcal{L}_{obs}$} &  \underline{0.122} & \underline{0.110}  & 0.074 & \underline{0.061}\\ 
{$\mathrm{Ours}$ w/o $\mathcal{L}_{sup}$}    &  \underline{0.105} &  \underline{0.095}          &0.069 &0.058 & {$\mathrm{Ours}$ w/o $\mathcal{L}_{sup}$} & 0.128 &0.119& \underline{0.071} &0.064\\ 
\rowcolor{pink!40}{Ours}  &  {\textbf{0.099}}  & \textbf{0.089}& \textbf{0.055} & \textbf{0.052} &{$\mathrm{Ours}$} & \textbf{0.120} & \textbf{0.102} & \textbf{0.070}& \textbf{0.057}\\ \bottomrule              
\end{tabular}
}
\caption{\textbf{Ablation studies with different loss functions on consistency.} We explored the effect of different loss on the consistency. }
\label{ala_loss}
\end{table}

\section{Experiments and Results}

We conducted evaluations of $\mathrm{M^{2}StyleGS}$ using two publicly available datasets: the LLFF dataset \cite{mildenhall2019local}, the Tanks \& Temples dataset \cite{knapitsch2017tanks}. For detailed experiment setup and more experiment results (limitation, user study, inference time, visual outcomes), \textbf{please check the Appendix.}

\subsection{Qualitative Results}

In employing reference images (\Cref{style_img}), M$^2$StyleGS demonstrates a more comprehensive style transfer outcomes. Compared to the baseline ConRF \cite{miao2024conrf}, which is remarkably blurry, especially in the scene ``Fern”, $\mathrm{M^{2}StyleGS}$ can perform finer clarity. Moreover, in contrast to StyleGaussian \cite{liu2024stylegaussian} and StyleRF \cite{liu2023stylerf}, which can maintain the integrity and fidelity of the original content but only achieves coarse color alignment, our method can maintain the integrity of the original content while effectuating a color-focused transfer. 

Regarding methods that utilize text prompts, CLIPNeRF \cite{wang2022clip} struggles to capture both the color richness and the deeper semantic information effectively. CLIPStyler \cite{kwon2022clipstyler}, a 2D transfer model, compromises too much content information as its excessive emphasis on textual descriptions detracts from the aesthetic quality of stylization, this issue is particularly noticeable in the “Trex” scene. Moreover, our method delivers more precise stylization outcomes compared to ConRF \cite{miao2024conrf}. This improvement is attributable to our modifications to the mapping module, which ConRF applies directly without alterations. 

\begin{figure}
\centering
\includegraphics[width=\linewidth]{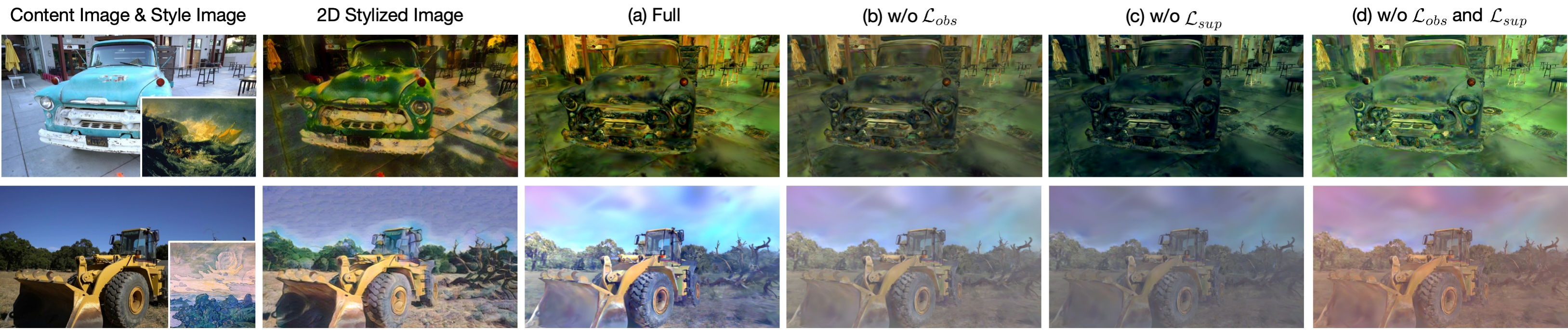}

\caption{\textbf{Ablation study with different loss functions.} (a): full system.  (b): stylization without $\mathcal{L}_{obs}$. (c): stylization without $\mathcal{L}_{sup}$. (d): stylization without $\mathcal{L}_{obs}$ and $\mathcal{L}_{sup}$. }
\label{loss}
\end{figure}

\subsection{Quantitative Results}

We compared our method with the SOTA methods in terms of multi-view consistency in \Cref{quan}. Following previous works \cite{liu2023stylerf,liu2024stylegaussian,miao2024conrf,miao2024ctnerf}, we wrap one view to another based on optical flow \cite{teed2020raft}, then computed masked RMSE ($\downarrow$) and LPIPS ($\downarrow$) \cite{zhang2018unreasonable} to measure the short- and the long-range consistency. 

In \Cref{quan}, among image-guided methods, compared with ConRF and StyleRF, the short-range LPIPS of $\mathrm{M^{2}StyleGS}$ is reduced by 32.92\% and 25.68\%. Note that for both long-range and short-range consistency, StyleGaussian achieves lower LPIPS and RMSE than other SOTA methods. This indicates that StyleGaussian can also produce novel scene views that maintain consistency, but its capability of color transfer still need improve.

For comparison with other text-guided methods, CLIPStyler exhibits a higher margin of LPIPS and RMSE than all SOTA methods. This discrepancy primarily stems from CLIPStyler lacking the utility and effect required for 3D scenes, leading to weaker multi-view consistency maintenance. CLIPNeRF maintains a reasonable level of consistency, and M$^2$styleGS significantly outperforms ConRF across all metrics, further demonstrating its superior transfer accuracy during different ranges of views.

\subsection{Ablation Studies}
To evaluate the effectiveness of subdivisive flow, we performed ablation experiments separately for $r$ round subdivisive flow. In the \Cref{abl}, we calculated the cosine similarity (SIM) ($\uparrow$) and the fréchet inception distance (FID) ($\downarrow$) between the original VGG feature ${X}_{s}^{v}$ and the CLIP feature $X_{s}^{c}$ in different subdivision. The statistics indicate that the original CLIP feature $X_{s}^{c}$ has no correlation with the VGG feature ${X}_{s}^{v}$ with the SIM score of -0.042. The presence of a mapping module gives a certain correlation between the mapped CLIP feature $\hat{X}_{s}^{c}$ and the VGG feature with FID decreased. As the subdivisive flow progresses, the color quality of the outcomes is significantly enhanced. A more precise alignment between the two features is achieved by rising 0.032, 0.083, and 0.137 in SIM among image transfer; declining 0.667, 0.721, 1.377 in FID among text transfer.

Combining \Cref{loss} and \Cref{ala_loss}, we present intuitive results that demonstrate the effectiveness of the introduced loss terms. \Cref{loss} (a) and (b) improve that the scene color distribution aligns more closely with that of the 2D priori with the intervention $\mathcal{L}_{obs}$. \Cref{loss} (a) and (c) elaborate that with the aid of $\mathcal{L}_{sup}$, the scene color distribution aligns more closely with that of the reference image. 
\Cref{ala_loss} illustrates that removing the auxiliary loss terms results in a slight increase in RMSE and LPIPS, which adversely affects the performance of the style transformation. This increase in RMSE and LPIPS highlights the critical role that the observation loss and suppression loss play in maintaining the multi-view consistency for high-quality multi-modality stylization.

\section{Conclusion}
In this work, we introduced M$^2$StyleGS, which leverages CLIP multi-modality knowledge to refine styles for instant 3D style transfer. M$^2$StyleGS employs a subdivisive flow module to facilitate feature matching and effectively addresses common issues observed in some SOTA methods, such as anomalous coloration or smoothed textures. We also propose two auxiliary loss functions to enhance the transferred visual effects. M$^2$StyleGS meets advanced user requirements without being limited to a single image style condition, offering the potential for various 3D applications.


\bibliography{egbib}

\end{document}